\documentclass[letterpaper]{article} 
\usepackage{aaai23}  
\usepackage{times}  
\usepackage{helvet}  
\usepackage{courier}  
\usepackage[hyphens]{url}  
\usepackage{graphicx} 
\usepackage{natbib}  
\usepackage{caption} 
\usepackage{algorithm}
\usepackage{algorithmic}

\usepackage[namelimits]{amsmath} 
\usepackage{amssymb}             
\usepackage{amsfonts}            
\usepackage{mathrsfs}            
\usepackage{graphicx} 
\usepackage{subfigure} 
\usepackage{xcolor}
\usepackage{makecell}
\usepackage{booktabs}
\usepackage{multirow}

\usepackage{xspace}
\newcommand{\method}{CoP\xspace}
\usepackage{newfloat}
\usepackage{listings}
\DeclareCaptionStyle{ruled}{labelfont=normalfont,labelsep=colon,strut=off} 
\floatstyle{ruled}
\newfloat{listing}{tb}{lst}{}
\floatname{listing}{Listing}
\title{CoP: Factual Inconsistency Detection by Controlling the Preference}
\author{
    Shuaijie She, 
    Xiang Geng, 
    Shujian Huang\thanks{Corresponding author}, 
    Jiajun Chen
}
\affiliations{
     National Key Laboratory for Novel Software Technology, Nanjing University\\
    
    \{shesj, gx\}@smail.nju.edu.cn, \{huangsj,chenjj\}@nju.edu.cn

}

\begin{document}

\maketitle

\begin{abstract}
Abstractive summarization is the process of generating a summary given a document as input. Although significant progress has been made, the factual inconsistency between the document and the generated summary still limits its practical applications. Previous work found that the probabilities assigned by the generation model reflect its preferences for the generated summary, including the preference for factual consistency, and the preference for the language or knowledge prior as well. To separate the preference for factual consistency, we propose an unsupervised framework named \method by controlling the preference of the generation model with the help of prompt. 
More specifically, the framework performs an extra inference step in which a text prompt is introduced as an additional input. 
In this way, another preference is described by the generation probability of this extra inference process. 
The difference between the above two preferences, i.e. the difference between the probabilities, could be used as measurements for detecting factual inconsistencies.
Interestingly, we found that with the properly designed prompt, our framework could evaluate specific preferences and serve as measurements for fine-grained categories of inconsistency, such as entity-related inconsistency, coreference-related inconsistency, etc.
Moreover, our framework could also be extended to the supervised setting to learn better prompt from the labeled data as well.
Experiments show that our framework achieves new SOTA results on three factual inconsistency detection tasks.

\end{abstract}
\section{Introduction}

Current abstractive summarization models~\cite{bart} can generate fluent summaries which are rich in information. However, over 70\% of the summaries generated by existing models contain factual inconsistency, which greatly damages the quality of the summary and limits the practical applications~\cite{faithfulness, frank}. 
The first step to solve this problem is to evaluate the factual consistency of the summary, in other words, to detect inconsistency~\cite{assess}.

Some research directly uses the preference of the probabilistic generation model, i.e. the generation probability, as an indicator for factor inconsistency~\cite{bartscore}.
But, since the generation model~\cite{bart} is usually pre-trained on large-scale corpora, the model would also have a preference for language or knowledge prior. 
To solve the problem, \citet{coco} introduces an extra inference with a partially masked document as input and uses the probability difference as the indicator for inconsistency.
However, since the masked document is inherently against the language prior, due to its nonfluency, the evaluation is still entangled by the preference of the model. Besides, it is also difficult to decide which word to mask in the document.

In this paper, we propose an unsupervised framework to elaborately \underline{co}ntrol the \underline{p}references with the help of prompt, namely \method.
\method employs a fluent text prompt, i.e. the summary itself, as the additional input for the extra inference step, which is easily to obtain and less relevant to the preference of the model. The resulting generation probability is conditioned on both the document and the prompt. 
So that we can use the differential probability between the two inference processes to detect factual inconsistency.

One advantage of the proposed framework is that the preference could be easily controlled by varying the prompt.
With proper designed prompts, \method could evaluate the specific preference and serve as measurements for fine-grained categories of inconsistencies, such as the entity-related inconsistency, the coreference-related inconsistency~\cite{frank}, etc.

Our framework not only works well in the unsupervised manner; its performance could be further improved by supervised training with prompt learning techniques~\cite{prefixtuing,gpttoo}.
Compared with previous work~\cite{dhc,Annotating} which fine-tune a large model, we could fine-tune a prompt vector with limited labeled data, which enjoys advantages in both effectiveness and efficiency.

Experiments are conducted on token-level and summary-level inconsistency detection and inconsistency category detection, where \method achieves remarkable improvements over several strong baselines in the unsupervised settings.
For detecting inconsistency categories, using specifically designed prompts brings further improvements without any additional training, revealing the flexibility of the prompt design and the great potential of our approach. 
With supervised training of the prompt, large improvements are achieved with only 0.02\% of the total parameters updated, showing an efficient strategy for the low-resource problem. In all the above experiments, our framework achieves better performance than the current state-of-the-art, to our best knowledge.\footnote{Code will be released at https://github.com/NJUNLP/CoP} 

\section{Preliminaries}
\subsection{Abstractive Summarization}
The essence of abstractive summarization is a conditional generation process: generating the target summary $Y=\{y_1,y_2,\dots,y_n\}$ with a given document $X=\{x_1,x_2,\dots,x_m\}$ as input.
In the summarization process, there is a abstractive summarization  model $M$ with parameter $\theta$. In each inference step $i$, the model will give the probability distribution of the current token $y_i$ based on the following 
 probability:
$ P(y_i | X,y_{<i};\theta)$,
where $y_{<i}=\{y_1,y_2,\dots,y_{i-1}\}$ denotes the prefix tokens~\cite{entfa,coco}.

\subsection{Factual Inconsistency Detection}
Given a document-summary pair  $(X, Y)$, the factual inconsistency detection task is to decide whether $C$ is supported by the document~\citep{faithfulness,dhc},
where $C$ can be defined in multiple granularities. 
In summary-level, $C$ is the whole summary, meaning that the task is to decide whether the whole summary is consistent with the given document. In this setting, a factual score $F$ is usually assigned to each summary.  
In a finer granularity,  $C$ could be each token in the summary, meaning that the task is to check token-level factual consistency, which assigns a factuality label for each token.
\begin{table}[ht]
\centering
\begin{tabular}{p{0.95\columnwidth}}
\toprule
\textbf{Document:}\\
\midrule
Mr Charney, who also founded the company, was ousted last year because of the employee complaints and amid accusations of misuse of company funds. In the San Francisco court filing, the board said it did not expect \dots  \\
\midrule
\textbf{Summary (with Token-Level Labels):}\\
\midrule
The former chief executive of the \underline{San} \underline{Francisco} \underline{court} , Charney , has been \underline{cleared} by a court in California.\\
\midrule
\textbf{Summary-Level Labels  : }0 (inconsistent)\\
\midrule
\textbf{Category Labels :} EntE, OutE\\
\bottomrule
\end{tabular}
\caption{\label{sample}
An example of factual inconsistency detection in multiple granularities (underlined words are inconsistent words in the token level; EntE and OutE are inconsistency categories, denoting Entity Error and Out of Article Error). 
}
\end{table}

Factual inconsistencies could also be related to different categories. Besides sentence-level and token-level labels, another task is to specify the detailed type $T$ of inconsistency. In this paper, we follow the typology definition of factual inconsistency from \citet{frank}. 
Table~\ref{sample} shows an example: the four underlined tokens, ``San Francisco court" and ``cleared'' are token-level inconsistent contents $C$.
The summary-level score $F$ indicates that the summary is inconsistent. Finally, the category label EntE represents the summary containing entity-related inconsistency.

\begin{figure}[th]
\centering
\includegraphics[scale=0.40]{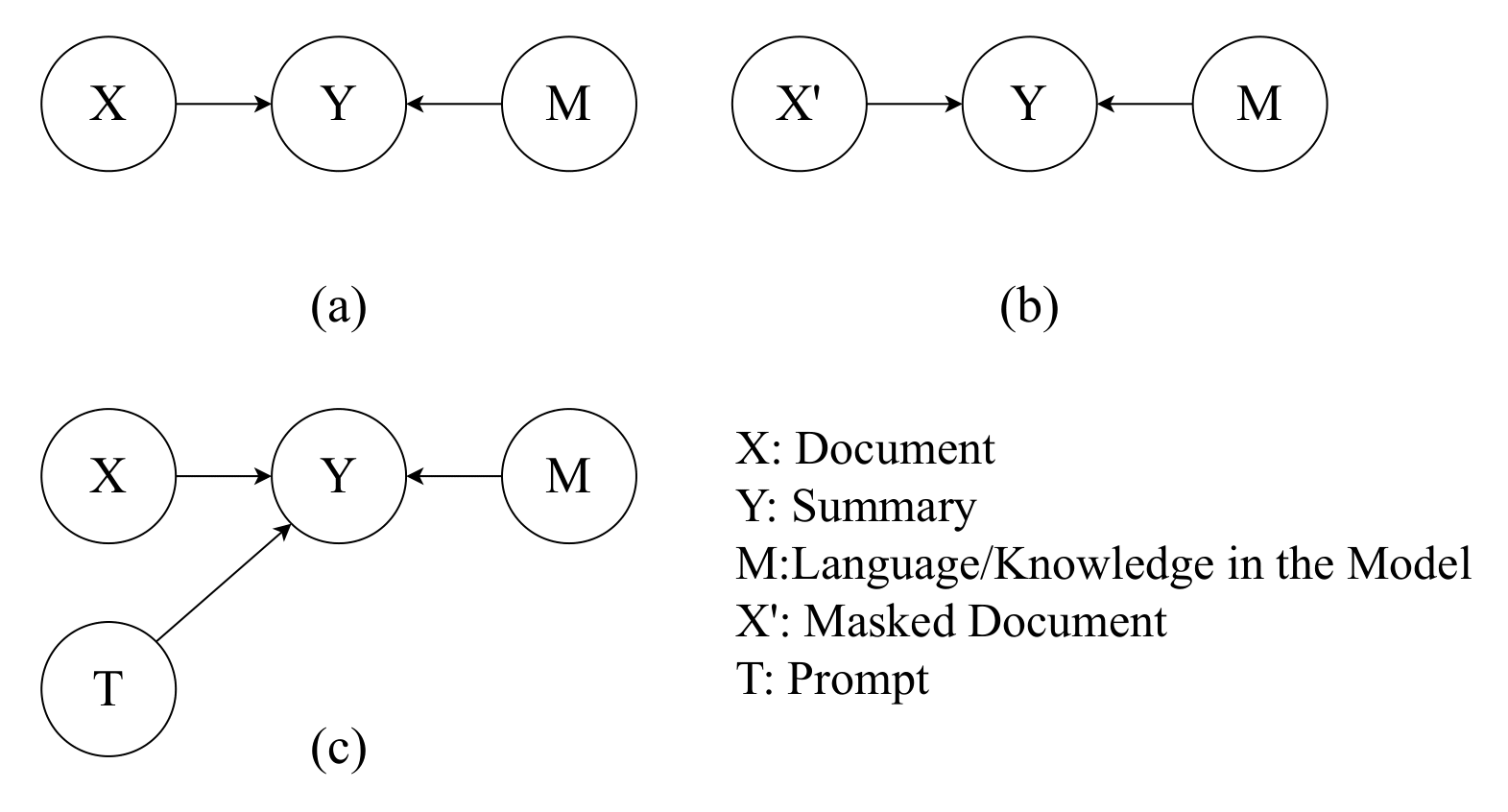}
\caption{Illustration of different inference processes: (a) the regular inference, where the generation of Y is determined by the document and the pre-trained model; (b) the inference in \citet{coco} with a partially masked document;  (c) our extra inference step with an additional prompt.}
\label{fig_casual_graph}
\end{figure}

\begin{figure*}[th]
\centering
\includegraphics[scale=0.65]{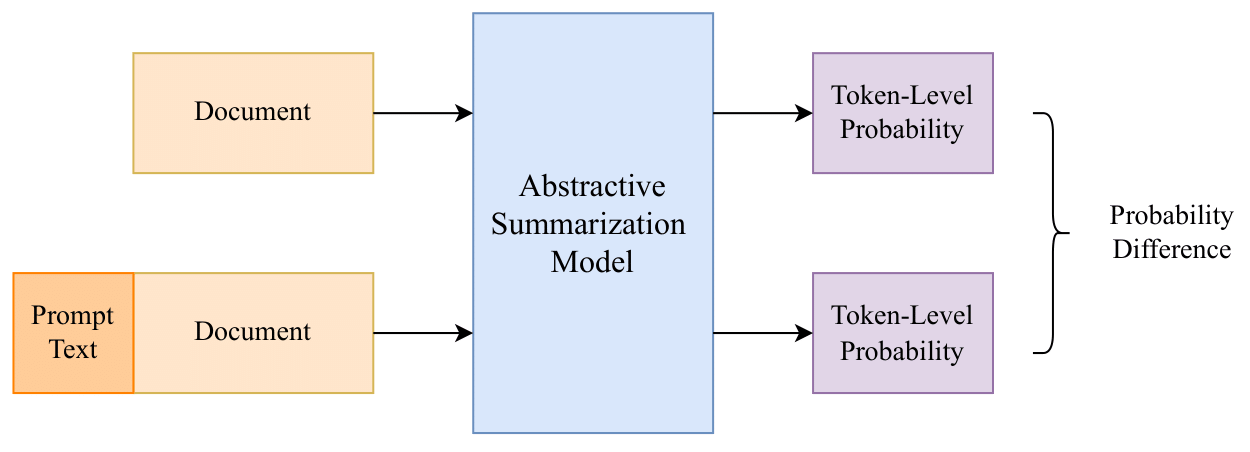}
\caption{Our framework of controlling the preference: \method}
\label{fig_prompt-tuing}
\end{figure*}

\section{Our Method}
\subsection{Motivation}\label{motivation}
The generation of a summarization involves at least the following two factors. The document $X$ provides the important source of factual information to support the consistent summary, leading to a preference for factually consistent summaries. On the other hand, the model $M$, usually pretrained on larger texts, provides the essential language prior knowledge to make the summary fluent in the generation process, i.e. preferring fluent summarization result. 
Therefore, the generation probability $p$ of each summary token is jointly decided by the model $M$ and source document $X$. The above causal relations are presented in Figure.~\ref{fig_casual_graph}(a).

The essence of the factual inconsistency detection is to evaluate how much the summary $Y$ is supported by the document $X$, that is to estimate the causal effect of $X \rightarrow Y$. Obviously, simply using the probability from the regular inference process, such as BARTScore~\cite{bartscore}, cannot discriminate the preferences from $X$ and $M$ and may lead to errors in detection. For example, some factually consistent but not fluent summaries will be misjudged as inconsistency.

The probability differential approach employs probability from an extra inference process to disentangle the preference from $X$ and $M$. \citet{coco} propose to introduce a partially masked document as the input for the extra inference~(Figure~\ref{fig_casual_graph}(b)). However, because the masked document is usually less fluent which also contradicts the language prior knowledge, 
the generation probability will still be entangled with $M$.
Moreover, it is difficult to decide which document token to mask precisely.

We propose a different extra inference process to obtain the differential probabilities. Instead of disturbing the original input X, we keep X as it is, but employ an extra prompt as the additional input (Figure~\ref{fig_casual_graph}(c)). The prompt could be fluent text, e.g. the summary itself, which does not affect the preference from $M$. So the differential probability would be mainly affect by the preference of $X$.
We present our framework \method, and its advantages in the rest of this section.


\subsection{Controlling the Preference by an Extra Inference with Prompt}\label{base}

Our framework consists of two inference processes (Figure~\ref{fig_prompt-tuing}). The first process is the same with regular inference, where the document is used as input and the decoder is forced to generate the given candidate summary using forced-decoding~\cite{bartscore}.
The log probability of generating each token in the candidate summary is denoted as $P_1$. 
The second inference process takes an text prompt $T$ together with the document $X$ as the input. With the same forced-decoding process, a second log probability is obtained, denoted as $P_2$. 

The choice of $T$ is flexible and may adapt to different scenarios. In the simplest case, we take the candidate summary $Y$ as the prompt (also called prompt text specifically). The intuition lies in how the prompt affects the generation probability. Intuitively, the consistent part in the summary is redundant to the generation model and thus will not bring large changes in probability. On the other hand, the inconsistent part in the summary will bring a more significant variation in its generation probability. In other words, the probability difference is caused by the model's preference for factual consistency, which filters out the irrelevant preferences such as fluency. In practice, we can calculate $P_{\text{diff}}$ of each token in the summary by following equation:
\begin{equation}
    P_{\text{diff}}(y_i) = P_2(y_i)-P_1(y_i)
\end{equation}

Tokens with higher $P_{\text{diff}}$ are more likely to be factual inconsistent with the document.
According to the certain applications, we can set different thresholds to control the proportion of predictions, and the tokens above the threshold are identified as inconsistent. For example, we can choose a relatively low threshold for inconsistent error correction task which higher recall is preferred.

Please note that in the area of prompt learning, prompt is the extra input added to the original input, which motivates the model to use its pre-trained abilities to accomplish specific tasks~\cite{promptsurvey}. Here we use prompt as an variant of input, which motivates the model to generate different probabilities. The method is simple and requires no additional training.

\subsection{Design Prompt for Fine-Grained Category}\label{ensemble}
Previous work~\cite{frank} defined a typology of the factual inconsistency, annotated datasets and analyzed the distribution of inconsistency category.
However, existing factuality assessment methods usually overlook these category annotations and only use the overall score for evaluation. 
We argue that the detailed category information of inconsistency will be helpful to analyze inconsistency tendency of existing models and provide guidance for future improvement.
According to~\citealp{frank}, EntE (Entity-related Error), CorefE (Coreference-related Error) and OutE (Not in Article Error) appear with a high frequency of 36\%, 10\% and 27\%, respectively. 
We try to address this issue and first consider the three most frequent inconsistency categories to illustrate how our framework works. Since OutE corresponds to errors that are not in the article, we can solve this inconsistent category with the candidate summary as prompt. So we also included it in the experiment as a comparison.

In addition to using our framework directly, we further design various prompts to control the preference of generation model. The base version of prompt text is the whole candidate summary that cover all inconsistent tokens in the summary. For detecting inconsistencies in each category, we could add fact information related to the specific category to improve the detection process. 
The intuition is similar: the consistent category-related information is redundant to the generation model, while the inconsistent category-related information will bring a more significant category-targeted probability variation.

 In this case, we can extract entities from the summary and append the entity token list after the original prompt text. If the probability of the summary is greatly affected by additional entity information, we can consider that this summary contains inconsistency of entity category. For CorefE, we can start with reference resolution of the summary and add these fact information: insert resolved reference token after the pronoun tokens. 
We believe that the addition of entities adds more entity information to the prompt text, bringing more entity-targeted probability changes. And the co-reference information can help model analyze and compare the co-reference facts in the abstract.

\begin{figure}[th]
\centering
\includegraphics[scale=0.65]{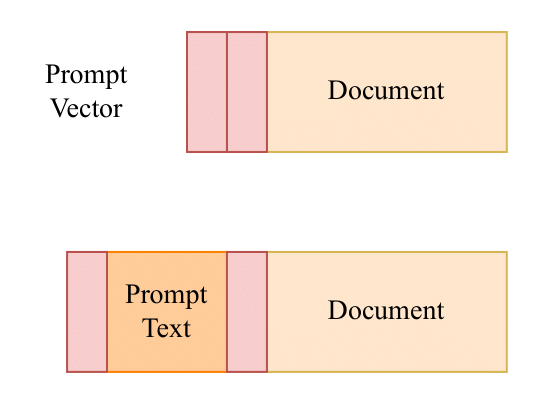}
\caption{Illustration of prompt vector (marked in red).}
\label{pve}
\end{figure}

At the moment we are still getting token-level scores, and category annotations are at the summary-level. The most straightforward way is to simply average all the token-level scores (all tokens are equally weighted). It's worth noting that our framework can finely check the factuality of each token in the summary, including the entity tokens and pronoun tokens. We can make better use of this information. We simply double the weights of entity tokens and pronoun tokens to help the model focus on the specific category.

\subsection{Learn From Labeled Data with Prompt Tuning}

Although our framework already works well in training-free style by exploiting the preference in summarization model, it is still possible to take advantage of the labeled data, which maybe extremely scarce. Thanks to the flexible structure of our framework, we can integrate the prompt tuning technique, which achieves great success in few-shot tasks. 

To learn a prompt text in discrete tokens is difficult, instead, we propose to a small-sized continuous task-specific vector (we call it \textbf{prompt vector} as opposed to \textbf{prompt text}).
We expect the prompt vector to help the model distinguish the prompt text and document, and guide the model to make a fine inferential comparison on factuality between the candidate summary and the document. More specifically, we add the same prompt vector before and after the prompt text in the extra inference step as shown in Figure~\ref{pve}. 
To keep the inference similar in the two processes, the same prompt vector are added to the regular inference process with no prompt text between them. To train the prompt vectors, we froze the whole summarization model and only update the small-scale prompt vector with the following loss function:
\begin{equation}
    \mathcal{L} = \sum_i P_{\text{diff}}(y_i) * label 
\end{equation}
$label$ is the token-level label of 1 and -1, indicating whether $y_i$ is consistent or inconsistent, respectively. Our loss function is designed succinctly for task purposes: further maximizing the $P_{\text{diff}}$ of inconsistent tokens and minimizing the $P_{\text{diff}}$ of consistent tokens. Minimizing the loss with labeled data strengthens the preference for factual consistent and provides a clearer discriminating boundary.

\begin{table*}[htbp]
\setlength{\tabcolsep}{3mm}
\centering
\begin{tabular}{lcccccccccccccl}\toprule
\textbf{Model} & \textbf{PtGen} & \textbf{TConvS2S}  & \textbf{TranS2S}  & \textbf{BERTS2S} & \textbf{Average F1} \\\midrule
Alignment-based *  & 38.92 & 37.94 & 34.47 & 35.81 & 36.78  \\
Overlap-based * & 57.22 & 54.25 & 53.79 & 55.13 & 55.09  \\
Synonym-based *  & 59.54 & 63.73 & 58.66 & 53.07 & 58.75  \\
DHC Model  & 64.72 & 69.37 & 63.88 & 56.49 & 63.61  \\
BARTSc-Ours * & 60.59 & 63.41 & 59.76 & 51.82 & 58.90 \\
\midrule
Ours Zero-Shot * & 63.08 & 68.61 & 63.90 & 58.58 & 63.54  \\
Ours Few-Shot & 68.27 & 70.31 & 66.03 & 60.19 & 66.20  \\
Ours Full-Shot & \textbf{71.06} & \textbf{72.91} & \textbf{67.64} & \textbf{65.67} & \textbf{69.32}  \\\bottomrule
\end{tabular}
\caption{\label{fine-grained result1}
F1 $(\times100)$ on each benchmark split, * denotes that this method is training-free
}
\end{table*}

\begin{table}[htbp]
\setlength{\tabcolsep}{3mm}
\centering
\begin{tabular}{lcccl}\toprule
\textbf{Model}  & \textbf{Corpus F1} \\\midrule 
DAE-Weak   & 59.10 \\
BARTSc-Ours * & 59.25 \\
EntFA & 60.23 \\
Ours Zero-Shot * & \textbf{63.72} 
\\\midrule 
DAE & 65.00 \\
Ours Few-Shot  & \textbf{66.56} \\
Ours Full-Shot & \textbf{69.61} \\\bottomrule
\end{tabular}
\caption{\label{fine-grained result2}
Corpus-Level F1 $(\times100)$ Performance, * denotes that this method is training-free
}
\end{table}
\section{Experiment Setup}

\subsection{Detect Inconsistency in Unsupervised Manner}
\textbf{Dataset:} \label{unsuper}
XSum Hallucination Annotations~\cite{faithfulness} sampled 500 document-summary pairs from the summarization dataset XSUM~\citep{xsum} and explored the fact consistency of summary generated by four popular models (PtGen, TConvS2S, TranS2S, BERTS2S). The annotator marks each token with 0 or 1 to indicate its factuality.
In addition to the above token-level dataset, we also take two popular summary-level datasets into consideration: QAGS~\citep{qags} and FRANK~\citep{frank} datasets. Both two datasets contain summary-level factuality annotations on CNNDM and XSUM datasets respectively.

\noindent\textbf{Implementation Details:} We take released BARTCNN\footnote{https://huggingface.co/facebook/bart-large-cnn} (BART~\cite{bart} fine tuned on CNNDM dataset) as the backbone of our framework. 
Our framework ranks tokens with the $P_{\text{diff}}$ as score and consider tokens with scores above the threshold as inconsistency. 
For a fair comparison, we follow the setting in DHC~\cite{dhc} by setting the threshold to predict a close proportion of inconsistency.

\subsection{Improve Performance with Efficient Prompt Tuning}
\textbf{Dataset:} 
Following previous work,  we use the token-level dataset as in Section~\ref{unsuper}. Intuitively, token-level inconsistency detection is more difficult and requires the fine-grained analysis capabilities of the model.

\noindent\textbf{Implementation Details:}
In full-shot setting, the dataset is split into three subsets: training (1200), validation (400), and test (400) sets which is similar to DAE~\cite{Annotating}. We cut the training set to 300 for the few-shot settings. The length of prompt vector is set to 40 and 5 for full-shot and few-shot respectively. The prompt tuning process uses AdamW optimizer with 1e-3 learning rate and the experiment are conducted on single TITAN-RTX GPU.

\subsection{Detailed Inconsistency Category Evaluation}
\textbf{Dataset:} 
In addition to the factual consistency scores at the summary level, FRANK~\citep{frank} dataset provides more detailed assessment information by showing the types of inconsistency in each summary. On the FRANKCNN set, the candidate summary is longer and there are richer and more comprehensive inconsistency types, so we use this dataset to verify the effectiveness of our method.

\noindent\textbf{Implementation Details:} For CorefE, we use SpanBERT~\citep{spanbert} to generate coreference targeted prompt text. According to the definition of CorefE, We filtered the summaries without pronouns. For EntE, we use Spacy\footnote{https://spacy.io/} to extract entity prompt text. The method of using the above strategy is called Ours Variant.

\section{Experiment Results and Comparison}
\subsection{Detect Inconsistency in Unsupervised Manner}
Following DAE and DHC, we report F1 on each data split in Table~\ref{fine-grained result1} and corpus-level F1 in Table~\ref{fine-grained result2}. From a train-free perspective, Ours Zero-Shot achieves impressive performance, increasing the F1 score by \textbf{4.64} compared to the best method BARTSc-Ours. It is a very intuitive demonstration of the effectiveness of the extra inference and prompt. 

 Compared with some methods training with large-scale pseudo-data, Our Zero-Shot still shows very competitive results, even outperforming DAE-Weak by \textbf{4.62} in corpus-level F1.
Moreover, our model has a steady improvement on each data subset, further proving the generalization of our model to handle different styles of candidate summaries generated by different summarization models well.

Table~\ref{summary-level} shows the pearson correlation with human score in summary-level inconsistency detection task. Our model achieve SOTA results in all four datasets.
It's worth noting that our model leads SOTA by a larger margin in two XSUM datasets: our method exceeded BARTScore by \textbf{3.98} and \textbf{5.34} on QAGSXSUM and FRANKXSUM respectively. This phenomenon may be related to the fact that XSUM is more abstractive and contains more noise which proves that the \method can better separate the unrelated preference for fluency and focus on the verification of fact consistency.

\subsection{Improve Performance with Efficient Prompt Tuning}\label{worse}
\begin{table*}[htbp]
\setlength{\tabcolsep}{3mm}
\centering
\begin{tabular}{lccccccccccccl}\toprule
\textbf{Model} & \textbf{QAGSCNN} & \textbf{QAGSXSUM}  & \textbf{FRANKCNN}  & \textbf{FRANKXSUM} \\\midrule
BERTScore  & 27.63 & 2.51 & 32.85 & 18.63 \\
QAGSScore & 54.53 & 17.49 & 30.45 & -5.847  \\
BARTScore  & 71.43 & 22.65 & 55.84 & 17.43   \\
CoCoScore & 58.84 & 24.08 & - & -  \\
Ours Zero-shot& 73.02 & 26.63 & 56.09 & 22.77  \\\bottomrule
\end{tabular}
\caption{\label{summary-level}
Summary-level pearson correlation$(\times100)$ between metrics and human judgments of factuality 
}
\end{table*}
We further validate the effectiveness of our designed prompt tuning in the token-level inconsistency detection task as shown in Table~\ref{fine-grained result1} and Table~\ref{fine-grained result2}. 
Using only 300 annotated data significantly improved the performance of our model to a new SOTA level, exceeding DAE, which uses 2K annotated data and DHC, which requires 960K pseudo-data. This suggests that \method can learn more efficiently from small amounts of data.

As the scale of annotated data increases, the performance of our model is further improved. When we train with the entire training set (1200), the corpus level F1 came to \textbf{69.61}, which is an improvement of 5.89 (9.24\%) over the impressive Our Zero-shot. 
Thanks to the effectiveness of the prompt tuning we designed, we were able to fully tap the potential of \method with the scarce data.
Another thing worth noting is that compared with DAE, which used more annotated data, we improved 4.61, showing a strong effectiveness.

\subsection{Detailed Inconsistency Category Evaluation}

\begin{table}[htbp]
 \centering
 \begin{tabular}{lcccccl}\toprule
    \textbf{Model}  & \textbf{Overall}& \textbf{EntE} & \textbf{OutE}\\\midrule
    BERTScore & 23.62 & 14.57 & 0.092 \\
QAGSScore & 30.15 & 21.40 & 12.19 \\
BARTScore & 55.98 & 32.58 & 18.56 \\
\midrule
Ours Base & 56.07 & 33.32 & 23.47 \\
Ours Variant & 57.66 & 34.66 & 26.09 
    \\\bottomrule
 \end{tabular}
 \caption{\label{EntE}Pearson correlation$(\times100)$ between metrics and human judgments of EntE}
 \label{tab:EntE}
\end{table}

\begin{table}[htbp]
 \centering
 \begin{tabular}{lcccccl}\toprule
 \textbf{Model}  & \textbf{Overall} & \textbf{CorefE} & \textbf{OutE}\\\midrule
 BERTScore & 31.33 & 24.65 & 15.86 \\
QAGSScore & 29.67 & 10.09 & 12.28\\
BARTScore & 56.38 & 29.29 & 34.76\\
\midrule
Ours Base & 59.76 & 34.77 & 37.89\\
Ours Variant & 60.57 & 35.47 & 36.55 \\\bottomrule
 \end{tabular}
 \caption{\label{CorefE} Pearson correlation$(\times100)$ between metrics and human judgments of CorefE}
 \label{tab:CorefE}
\end{table}

The experimental results of Table~\ref{CorefE} and Table~\ref{EntE} show that our baseline method is not only good at detecting fine-grained fact inconsistency but also good at detecting certain types of factual inconsistency errors. 
When Ours Base has exceeded the previous work, Ours Variants can even further improve the pearson correlation with human annotation in different categories of inconsistency. It is surprising enough to note that we did not have additional training to achieve this, but with the flexible design of the prompt. 

What is more surprising is that when the designed prompt improves the evaluation of specific categories, it also affects the pearson correlation of Overall and OutE. We think this is due to: (1) EntE is a fairly frequent inconsistency. When we improve the detection of this category, the model's ability to identify overall inconsistencies will also be improved. (2) There is some relation between inconsistency categories, such as between EntE and OutE. If the model fails to understand entities in document, it is also more likely to generate OutE. That's why a prompt for one category can help detect other inconsistency categories.

\section{Analysis}
\subsection{Case Study}

\begin{table}[ht]
\centering
\begin{tabular}{p{0.95\columnwidth}}
\toprule
\textbf{Document1:} Trevor Deely, 22, was last seen walking home from a Christmas party in December 2000. A search of a site in Chapelizod in Dublin started \dots \\
\textbf{Summary1:} A search is under way for a man missing in the republic of Ireland in the republic of Ireland.  \\
\textbf{CoCo:}  -4.45 \\
\textbf{Ours:}  -2.28 \\
\textbf{Human Annotation:} consistent \\
\midrule
\textbf{Document2:} The win means Egypt play Ghana on Wednesday with top spot of Group D still at stake, while Uganda are out of their first tournament since 1978. \\
\textbf{Summary2:} Uganda was knocked out of the Africa Cup of Nations as they were held to a \underline{goalless} \underline{draw} \underline{by} \underline{Uganda} at the Africa Cup of Nations.\\
\textbf{CoCo:}  -2.87 \\
\textbf{Ours:}  -4.12 \\
\textbf{Human Annotation:} inconsistent \\
\bottomrule
\end{tabular}
\caption{\label{casestudy}
Comparison of CoCo and Ours (\method). Higher score indicates that the model considers the summary to be more consistent. (underlined words are inconsistent)
}
\end{table}
We sampled two examples in Table~\ref{casestudy} and compared the assessment results of CoCo and our proposed \method. For the Summary1 , it is factually consistent with redundant generation which is a common problem in abstractive summary. 
If the evaluation model has difficulty separating fluency preferences, redundant spans that reduce the fluency will mislead the model into thinking that the summary is inconsistent. Obviously, our model is better to separate preferences and accurately judge that the summary is more consistent.

Another example, the sentence is quite fluent, but it has a common factual inconsistency: Entity Error. Such errors are frequent and subtle, making discrimination difficult. CoCo gives Summary2 a even higher score than Summary1 which is more consistent. On the contrary, \method shows better ability in evaluation factual consistency.

\subsection{Robustness on Different Backbones}
\begin{table}[htbp]
\setlength{\tabcolsep}{3mm}
\centering
\begin{tabular}{lccccccccccccl}\toprule
\textbf{Model} & \textbf{BART} & \textbf{PEGA}  & \textbf{BRIO}  & \textbf{T5}  \\\midrule
BARTSc  & 71.42 & 72.10 & 56.03 & 43.82\\
CoCo  & 58.84 & 58.96 & - & 56.13 \\ 
Ours & 73.01 & 74.01 & 65.97 & 69.29 \\ \bottomrule
\end{tabular}
\caption{\label{diffentbackbone}
Summary-level pearson correlation$(\times100)$ on QAGSCNN of \method and baselines on different backbones.
}
\end{table}
We tested our method based on various backbones on QAGSCNN dataset. We choose BART~\cite{bart}, PEGASUS~\cite{pegasus}, BRIO~\cite{brio}, T5~\cite{t5} and the experimental results are shown in Table~\ref{diffentbackbone}. \method achieves stable and impressive improvements on different backbones against strong baselines.

\subsection{Flexible Length of Prompt Vector}
\begin{figure}[ht]
\centering
\includegraphics[scale=0.30]{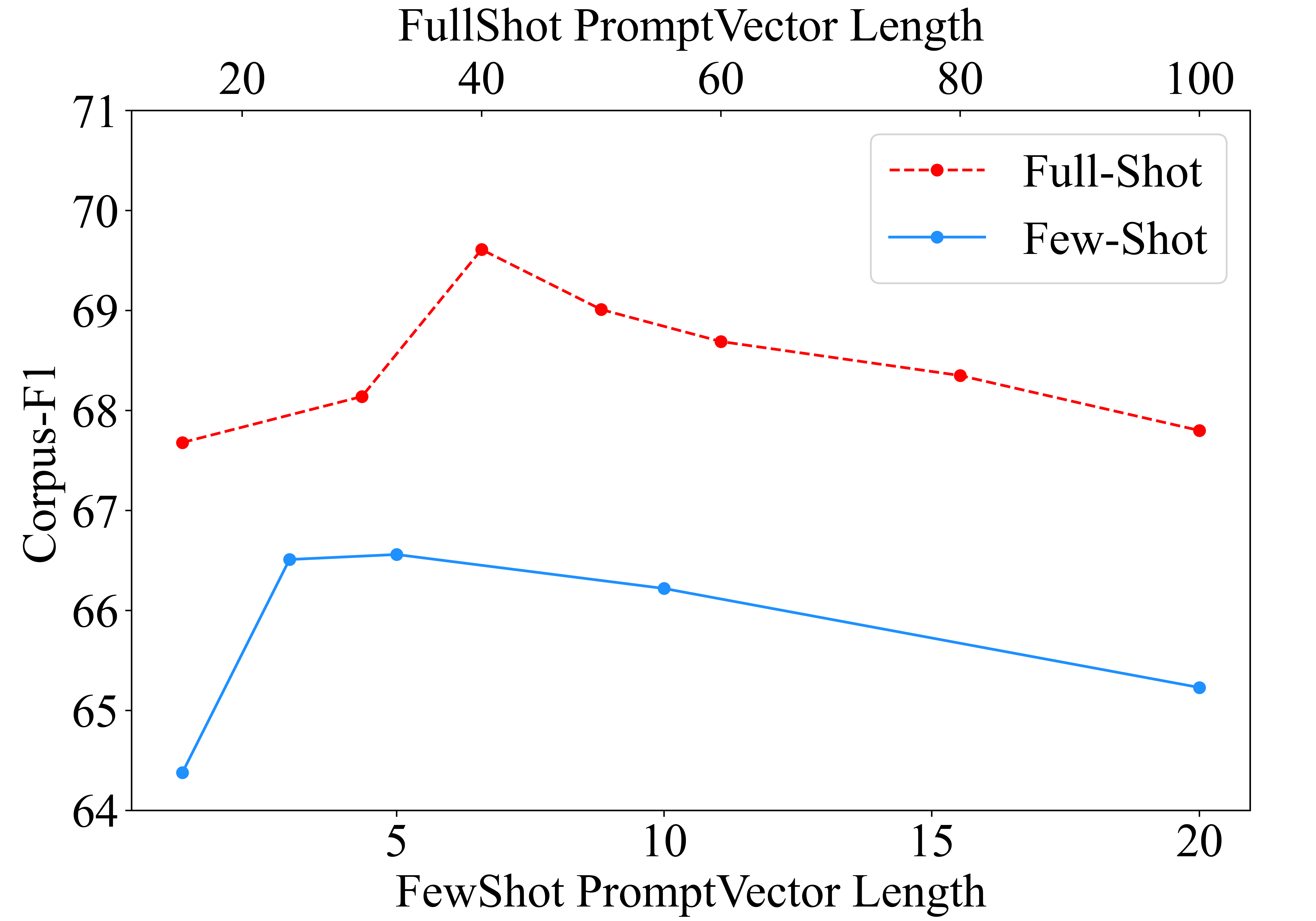}
\caption{Prompt vector length and corpus-level F1.  Two X-axis corresponding to two training settings.}
\label{prompt_length}
\end{figure}

Since we are the first to propose this novel training schema in factual consistency task, we analyzed the important properties of prompt vector: prompt length. 
As the length of the prompt vector increases, it means that there are more trainable parameters and the model will have more expressive ability. The Figure~\ref{prompt_length} shows that performance increases as the length of prompt vector increases up to a threshold. We also observed the slight performance degradation above the threshold. We think this is mainly because in this low-resource and noisy dataset, the model suffers from a risk of overfitting to the noise in the data. Fortunately, our adjustable training parameters help us deal with scenarios with different size of data, while the previous work can only fine tune from a fixed-size PLM. 
Moreover, even with an unreasonable prompt length (such as 100), our framework still outperforms the previous SOTA model which further demonstrates the robustness of our framework.

\begin{table}[htbp]
\setlength{\tabcolsep}{2mm}
\centering
\begin{tabular}{lccl} \toprule
\textbf{Model} & \textbf{Param Size}\\\midrule
DAE &  109M \\
DHC &  330M \\
EntFA &  406M \\
\midrule
Ours Zero-Shot &  0M \\
Ours Few-Shot &  0.01M \\
Ours Full-Shot &  0.08M \\
\bottomrule
\end{tabular}
\caption{\label{Models Trainable ParamsSize} Trainable parameter size of different methods.}
\end{table}
\subsection{Significantly Fewer Parameter for Efficiency}
The number of trainable parameters greatly affects the training speed and the size of needed GPU memory. 
Moreover, for the low resource task as factual consistency evaluation, previous methods have to construct large-scale pseudo-data for training. A large amount of pseudo-data further increases the cost of training and brings an irreparable performance gap compared to real data~\cite{Annotating}. We compare trainable parameter size between our framework and other methods in Table~\ref{Models Trainable ParamsSize}. The results of the comparison show that we only use \textbf{0.02\%} of the parameters to exceed the results of previous work which demonstrate that our framework is quite parameter-efficient.

\subsection{Prompt Tuning Contributes to Clearer Decision Boundaries}

\begin{figure}[htbp]
\centering
\includegraphics[scale=0.32]{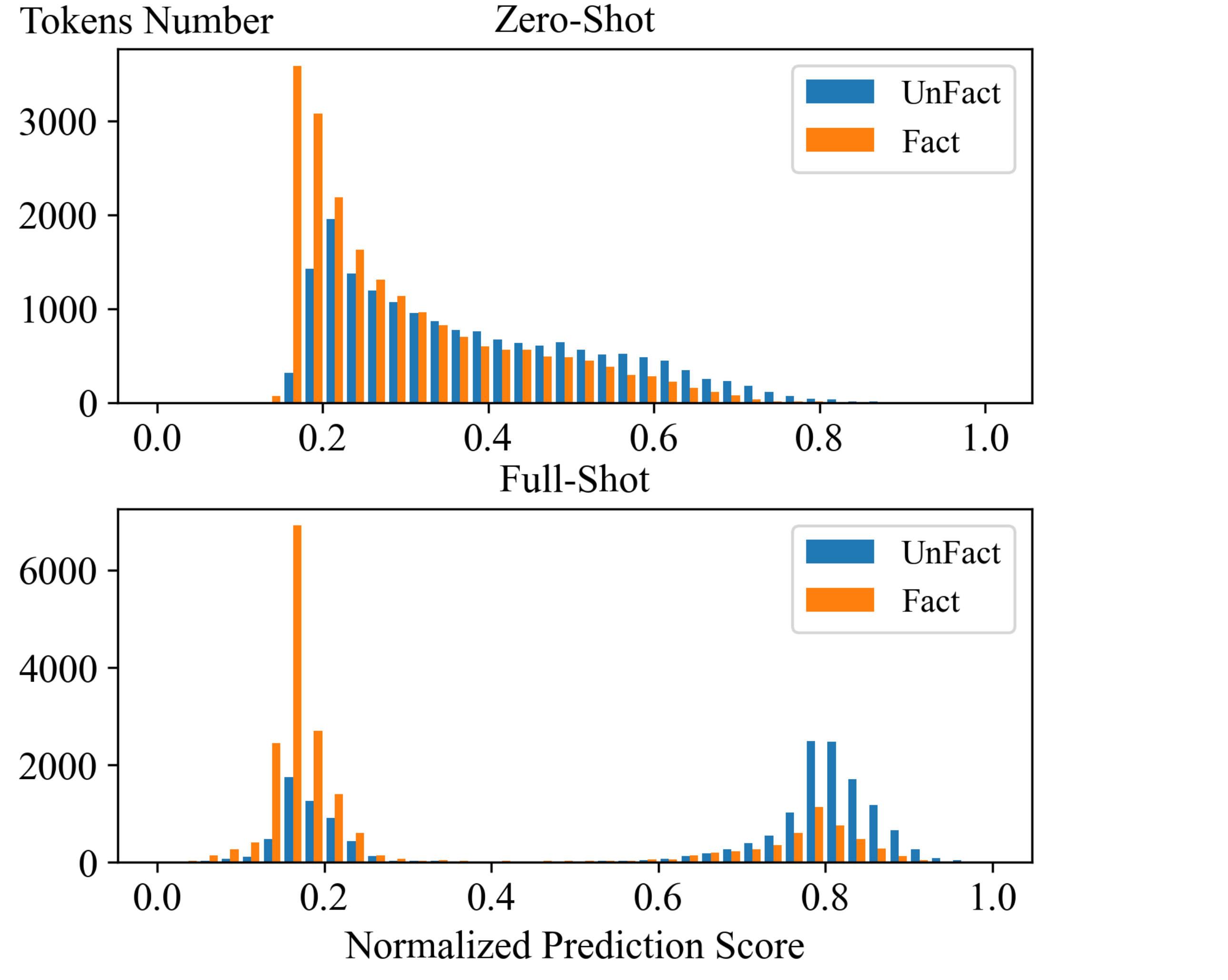}
\caption{Normalized score distribution. Higher score denotes \method thinks the token is more inconsistent.}
\label{distributions}
\end{figure}
The distribution of token-level score is visualized in Figure~\ref{distributions}. For comparison, we normalized the scores ($P_{\text{diff}}$).
The certain difference in distribution between factual and unfactual tokens explains why \method is valid under zero-shot. 
After enhancement by prompt tuning, the distribution presents a better decision boundary, thus distinguishing factual and unfactual tokens better. 
The distribution difference is a clear demonstration of the feasibility of our framework. 
\section{Related Work}
Previous work~\cite{frank,faithfulness} show that the common n-gram based evaluation toolkits ROUGE~\cite{rouge}, BLEU~\cite{bleu} can not perform the fact consistency task well. BERTScore~\cite{bertscore}, though, takes advantage of the contextual information and pre-trained model to improve the performance. However, its correlation with human judgments is still unsatisfactory.

Therefore, a series of work on the assessment of fact consistency has been proposed. Some work extracts facts for matching, such as QAGS~\cite{qags} using QA and QG to extract facts from summaries and documents. Some work like FactCC~\cite{factcc} directly translates factual consistency judgment into a sentence classification. Recently more and more work pay attention to the generation process. BARTScore~\cite{bartscore} evaluates consistency with generation probability. And CoCo~\cite{coco} models consistency problem by counterfactual and uses mask operations to estimate causal effects. These works mainly focus on coarse-grained at the summary level to measure factual consistency. 

In terms of fine-grained factual consistency assessment, DHC~\cite{dhc} constructs large-scale pseudo-data and trains a token-level classifier with Roberta~\cite{roberta}. ~\citet{Annotating} full tuned Electra~\cite{electra} on pseudo-data and small batches of real data. Experimental results show that the huge difference in distribution between pseudo-data and real data greatly limits the effectiveness of current models which encourages us to explore a novel training scheme.

\section{Conclusion}
In this paper, we propose \method that detects factual inconsistency by controlling the preference with the help of prompt. By separating irrelevant preference, \method can accurately measure factual inconsistency without training. Moreover, \method could evaluate the specific preference and detect detailed inconsistency categories without training. We also explore to improve the performance by prompt tuning with labeled data which achieved both efficiency and effectiveness. Experimental results on token-level and
summary-level factual inconsistency detection and detailed
inconsistency category detection demonstrates the effectiveness of our work.

\section{Acknowledgments}
We would like to thank the anonymous reviewers for their insightful comments. Shujian Huang is the corresponding author. This work is supported by National Science Foundation of China (No. 62176120), the Liaoning Provincial Research Foundation for Basic Research (No. 2022-KF-26-02).

\bibliography{aaai23}

\begin{thebibliography}{24}
\providecommand{\natexlab}[1]{#1}

\bibitem[{Cao, Dong, and Cheung(2021)}]{entfa}
Cao, M.; Dong, Y.; and Cheung, J. C.~K. 2021.
\newblock Inspecting the Factuality of Hallucinated Entities in Abstractive
  Summarization.
\newblock \emph{CoRR}, abs/2109.09784.

\bibitem[{Clark et~al.(2020)Clark, Luong, Le, and Manning}]{electra}
Clark, K.; Luong, M.; Le, Q.~V.; and Manning, C.~D. 2020.
\newblock {ELECTRA:} Pre-training Text Encoders as Discriminators Rather Than
  Generators.
\newblock \emph{CoRR}, abs/2003.10555.

\bibitem[{Goodrich et~al.(2019)Goodrich, Rao, Liu, and Saleh}]{assess}
Goodrich, B.; Rao, V.; Liu, P.~J.; and Saleh, M. 2019.
\newblock Assessing the factual accuracy of generated text.
\newblock In \emph{proceedings of the 25th ACM SIGKDD international conference
  on knowledge discovery \& data mining}, 166--175.

\bibitem[{Goyal and Durrett(2021)}]{Annotating}
Goyal, T.; and Durrett, G. 2021.
\newblock Annotating and Modeling Fine-grained Factuality in Summarization.
\newblock \emph{CoRR}, abs/2104.04302.

\bibitem[{Joshi et~al.(2019)Joshi, Chen, Liu, Weld, Zettlemoyer, and
  Levy}]{spanbert}
Joshi, M.; Chen, D.; Liu, Y.; Weld, D.~S.; Zettlemoyer, L.; and Levy, O. 2019.
\newblock SpanBERT: Improving Pre-training by Representing and Predicting
  Spans.
\newblock \emph{CoRR}, abs/1907.10529.

\bibitem[{Kryscinski et~al.(2020)Kryscinski, McCann, Xiong, and
  Socher}]{factcc}
Kryscinski, W.; McCann, B.; Xiong, C.; and Socher, R. 2020.
\newblock Evaluating the Factual Consistency of Abstractive Text Summarization.
\newblock In \emph{Proceedings of the 2020 Conference on Empirical Methods in
  Natural Language Processing (EMNLP)}, 9332--9346. Online: Association for
  Computational Linguistics.

\bibitem[{Lewis et~al.(2020)Lewis, Liu, Goyal, Ghazvininejad, Mohamed, Levy,
  Stoyanov, and Zettlemoyer}]{bart}
Lewis, M.; Liu, Y.; Goyal, N.; Ghazvininejad, M.; Mohamed, A.; Levy, O.;
  Stoyanov, V.; and Zettlemoyer, L. 2020.
\newblock {BART}: Denoising Sequence-to-Sequence Pre-training for Natural
  Language Generation, Translation, and Comprehension.
\newblock In \emph{Proceedings of the 58th Annual Meeting of the Association
  for Computational Linguistics}, 7871--7880. Online: Association for
  Computational Linguistics.

\bibitem[{Li and Liang(2021)}]{prefixtuing}
Li, X.~L.; and Liang, P. 2021.
\newblock Prefix-Tuning: Optimizing Continuous Prompts for Generation.
\newblock \emph{CoRR}, abs/2101.00190.

\bibitem[{Lin(2004)}]{rouge}
Lin, C.-Y. 2004.
\newblock Rouge: A package for automatic evaluation of summaries.
\newblock In \emph{Text summarization branches out}, 74--81.

\bibitem[{Liu et~al.(2021{\natexlab{a}})Liu, Yuan, Fu, Jiang, Hayashi, and
  Neubig}]{promptsurvey}
Liu, P.; Yuan, W.; Fu, J.; Jiang, Z.; Hayashi, H.; and Neubig, G.
  2021{\natexlab{a}}.
\newblock Pre-train, Prompt, and Predict: {A} Systematic Survey of Prompting
  Methods in Natural Language Processing.
\newblock \emph{CoRR}, abs/2107.13586.

\bibitem[{Liu et~al.(2021{\natexlab{b}})Liu, Zheng, Du, Ding, Qian, Yang, and
  Tang}]{gpttoo}
Liu, X.; Zheng, Y.; Du, Z.; Ding, M.; Qian, Y.; Yang, Z.; and Tang, J.
  2021{\natexlab{b}}.
\newblock {GPT} Understands, Too.
\newblock \emph{CoRR}, abs/2103.10385.

\bibitem[{Liu et~al.(2022)Liu, Liu, Radev, and Neubig}]{brio}
Liu, Y.; Liu, P.; Radev, D.; and Neubig, G. 2022.
\newblock BRIO: Bringing Order to Abstractive Summarization.
\newblock \emph{arXiv preprint arXiv:2203.16804}.

\bibitem[{Liu et~al.(2019)Liu, Ott, Goyal, Du, Joshi, Chen, Levy, Lewis,
  Zettlemoyer, and Stoyanov}]{roberta}
Liu, Y.; Ott, M.; Goyal, N.; Du, J.; Joshi, M.; Chen, D.; Levy, O.; Lewis, M.;
  Zettlemoyer, L.; and Stoyanov, V. 2019.
\newblock RoBERTa: {A} Robustly Optimized {BERT} Pretraining Approach.
\newblock \emph{CoRR}, abs/1907.11692.

\bibitem[{Maynez et~al.(2020)Maynez, Narayan, Bohnet, and
  McDonald}]{faithfulness}
Maynez, J.; Narayan, S.; Bohnet, B.; and McDonald, R. 2020.
\newblock On Faithfulness and Factuality in Abstractive Summarization.
\newblock In \emph{Proceedings of the 58th Annual Meeting of the Association
  for Computational Linguistics}, 1906--1919. Online: Association for
  Computational Linguistics.

\bibitem[{Narayan, Cohen, and Lapata(2018)}]{xsum}
Narayan, S.; Cohen, S.~B.; and Lapata, M. 2018.
\newblock Don{'}t Give Me the Details, Just the Summary! Topic-Aware
  Convolutional Neural Networks for Extreme Summarization.
\newblock In \emph{Proceedings of the 2018 Conference on Empirical Methods in
  Natural Language Processing}, 1797--1807. Brussels, Belgium: Association for
  Computational Linguistics.

\bibitem[{Pagnoni, Balachandran, and Tsvetkov(2021)}]{frank}
Pagnoni, A.; Balachandran, V.; and Tsvetkov, Y. 2021.
\newblock Understanding factuality in abstractive summarization with FRANK: A
  benchmark for factuality metrics.
\newblock \emph{arXiv preprint arXiv:2104.13346}.

\bibitem[{Papineni et~al.(2002)Papineni, Roukos, Ward, and Zhu}]{bleu}
Papineni, K.; Roukos, S.; Ward, T.; and Zhu, W.-J. 2002.
\newblock {B}leu: a Method for Automatic Evaluation of Machine Translation.
\newblock In \emph{Proceedings of the 40th Annual Meeting of the Association
  for Computational Linguistics}, 311--318. Philadelphia, Pennsylvania, USA:
  Association for Computational Linguistics.

\bibitem[{Raffel et~al.(2020)Raffel, Shazeer, Roberts, Lee, Narang, Matena,
  Zhou, Li, Liu et~al.}]{t5}
Raffel, C.; Shazeer, N.; Roberts, A.; Lee, K.; Narang, S.; Matena, M.; Zhou,
  Y.; Li, W.; Liu, P.~J.; et~al. 2020.
\newblock Exploring the limits of transfer learning with a unified text-to-text
  transformer.
\newblock \emph{J. Mach. Learn. Res.}, 21(140): 1--67.

\bibitem[{Wang, Cho, and Lewis(2020)}]{qags}
Wang, A.; Cho, K.; and Lewis, M. 2020.
\newblock Asking and Answering Questions to Evaluate the Factual Consistency of
  Summaries.
\newblock \emph{CoRR}, abs/2004.04228.

\bibitem[{Xie et~al.(2021)Xie, Sun, Deng, Li, and Ding}]{coco}
Xie, Y.; Sun, F.; Deng, Y.; Li, Y.; and Ding, B. 2021.
\newblock Factual Consistency Evaluation for Text Summarization via
  Counterfactual Estimation.
\newblock \emph{CoRR}, abs/2108.13134.

\bibitem[{Yuan, Neubig, and Liu(2021)}]{bartscore}
Yuan, W.; Neubig, G.; and Liu, P. 2021.
\newblock Bartscore: Evaluating generated text as text generation.
\newblock \emph{Advances in Neural Information Processing Systems}, 34.

\bibitem[{Zhang et~al.(2020)Zhang, Zhao, Saleh, and Liu}]{pegasus}
Zhang, J.; Zhao, Y.; Saleh, M.; and Liu, P. 2020.
\newblock Pegasus: Pre-training with extracted gap-sentences for abstractive
  summarization.
\newblock In \emph{International Conference on Machine Learning}, 11328--11339.
  PMLR.

\bibitem[{Zhang et~al.(2019)Zhang, Kishore, Wu, Weinberger, and
  Artzi}]{bertscore}
Zhang, T.; Kishore, V.; Wu, F.; Weinberger, K.~Q.; and Artzi, Y. 2019.
\newblock BERTScore: Evaluating Text Generation with {BERT}.
\newblock \emph{CoRR}, abs/1904.09675.

\bibitem[{Zhou et~al.(2020)Zhou, Gu, Diab, Guzman, Zettlemoyer, and
  Ghazvininejad}]{dhc}
Zhou, C.; Gu, J.; Diab, M.~T.; Guzman, P.; Zettlemoyer, L.; and Ghazvininejad,
  M. 2020.
\newblock Detecting Hallucinated Content in Conditional Neural Sequence
  Generation.
\newblock \emph{CoRR}, abs/2011.02593.

\end{thebibliography}

\end{document}